# TDPNavigator-Placer: Thermal- and Wirelength-Aware Chiplet Placement in 2.5D Systems Through Multi-Agent Reinforcement Learning

Yubo Hou[1], Furen Zhuang[1], Partha Pratim Kundu[1], Sezin Ata Kircali[1], Jie Wang[1], Mihai Dragos Rotaru[2], Dutta Rahul[2]
Ashish James[1]

[1] Institute for Infocomm Research (I²R), Agency for Science, Technology and Research (A*STAR), Singapore
[2] Institute of Microelectronics (IME), Agency for Science, Technology and Research (A*STAR), Singapore
Hou_Yubo@a-star.edu.sg

*Abstract*— The rapid growth of electronics has accelerated the adoption of 2.5D integrated circuits, where effective automated chiplet placement is essential as systems scale to larger and more heterogeneous chiplet assemblies. Existing placement methods typically focus on minimizing wirelength or transforming multi-objective optimization into a single objective through weighted sum, which limits their ability to handle competing design requirements. Wirelength reduction and thermal management are inherently conflicting objectives, making prior approaches inadequate for practical deployment. To address this challenge, we propose TDPNavigator-Placer, a novel multi-agent reinforcement learning framework that dynamically optimizes placement based on chiplet's thermal design power (TDP). This approach explicitly assigns these inherently conflicting objectives to specialized agents, each operating under distinct reward mechanisms and environmental constraints within a unified placement paradigm. Experimental results demonstrate that TDPNavigator-Placer delivers a significantly improved Pareto front over state-of-the-art methods, enabling more balanced trade-offs between wirelength and thermal performance.

## I. Introduction

In recent years, there has been growing demand for cost-effective, scalable chips across various markets such as processors, automotive electronics, and artificial intelligence [1]. However, as the pace of advanced technology node development slows, the design costs of System-on-Chip (SoC) solutions continue to rise. Against this backdrop, 2.5D integration is gaining recognition as an approach for developing cost-effective, large-scale chip systems and is being increasingly explored. It offers several advantages over traditional SoCs [2]. First, it helps reduce costs in both design and manufacturing stages while improving yield. Second, 2.5D integrated circuits (ICs) enable seamless heterogeneous integration of technologies and nodes within a single package, known as System-in-Package (SiP). Finally, it facilitates the reuse of chiplets, paving the way for more sustainable and complex systems.

Efficient development of versatile large-scale 2.5D-IC systems requires composable and dedicated design automation tools [3]. Among these, we focus on the critical challenge of optimal chiplet placement for performance maximization [4]. Prior research falls into three categories: Simulated Annealing (SA)-based [5, 6], enumeration-based [7], and reinforcement learning (RL)-based [8] approaches. The first category employs diverse layout representations including floorplans [9] and hierarchical B*-trees [10]. While capable of handling multiple performance metrics beyond wirelength, SA methods typically suffer from long runtime and suboptimal solution quality. In contrast, enumeration-based approaches can leverage pruning algorithms (e.g. branch-and-bound) and parallelization techniques to obtain superior solutions for layouts of limited size. For the last category, reinforcement learning agents place chiplets sequentially according to reward functions.

However, prior approaches have either optimized solely for wirelength or transformed multi-objective optimization into a single objective using techniques like weighted sum, min-max, distance functions etc.. Crucially, wirelength minimization and thermal management represent conflicting objectives, rendering previous methods ineffective. This leads to high power density and increases the risk of thermal failure in large-scale systems [11]. This challenge remains unresolved, particularly due to persistent high-temperature issues and the lack of comprehensive thermal-aware design exploration.

To address this challenge, we propose TDPNavigator-Placer, a novel multi-agent reinforcement learning (MARL) framework that dynamically optimizes chiplet placement based on their thermal design power (TDP). Unlike single-agent RL approaches that struggle with trade-offs between thermal and wirelength objectives, TDPNavigator-Placer explicitly delegates these inherently conflicting goals to specialized agents. Each agent operates under distinct reward mechanisms and environmental constraints within a unified placement paradigm, enabling focused optimization through explicit task decomposition. This separation is motivated by the observation that high-TDP chiplets are the primary drivers of thermal optimization challenges, while low-TDP chiplets have negligible impact on thermal performance. Our contributions are summarized as follows:

- We propose a multi-agent reinforcement learning framework explicitly designed to resolve the inherent conflict between wirelength minimization and thermal management in 2.5D IC placement.

- We proposes a novel "TDP navigator" module that assigns chiplets to specialized agents (wirelength-agent or thermal-agent) based on their TDP.

- We employs distinct agents with dedicated reward mechanisms (wirelength reward, thermal reward) and constraints (wire mask, thermal mask), operating under a unified placement paradigm with shared view mask, position mask and rotation position mask.

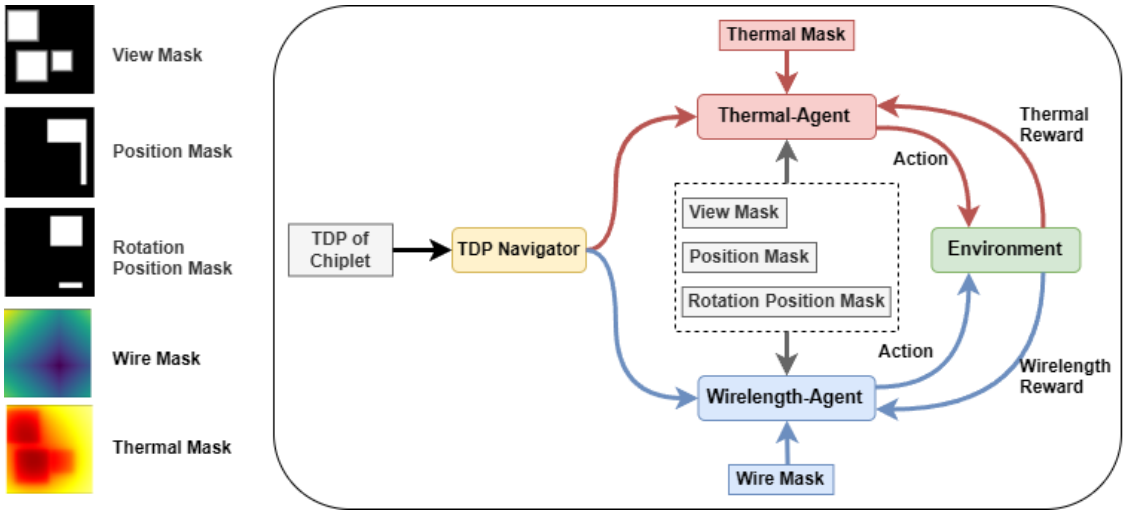

Fig. 1. Overview of TDPNavigator-Placer

- We evaluate our proposed method on various IC designs, demonstrating that it outperforms baseline methods by optimizing conflicting objectives — simultaneously reducing both temperature and wirelength.

## II. RELATED WORK

Existing mainstream chip placement methods can be broadly categorized into three types: analytical methods, black-box optimization (BBO), and learning-based approaches. Given that macros and chiplets are both coarse-grained modules requiring constrained placement, macro placement techniques provide an important theoretical and practical foundation for emerging chiplet optimization.

In analytical placement methods, GPU-accelerated nonlinear placement for designs containing large numbers of macros has become a popular research topic in recent years. For example, DREAMPlace [13] reformulates nonlinear placement as a neural network training task optimized using standard gradient descent. By leveraging GPU parallelism, this approach not only achieves significant computational speedups but also delivers the highest placement quality among current analytical methods.

BBO methods have been applied in placement for decades. Early approaches like SP [14] and B*-tree [10] suffered from poor scalability due to their reliance on rectangle-packing modelling. In contrast, recent research has focused more on optimizing the search space. For instance, AutoDMP [15] combines Bayesian optimization with DREAMPlace to efficiently explore the configuration space. WireMask-BBO [16] proposes a wire-mask-guided greedy genotype-phenotype mapping strategy that can be embedded into any BBO framework, consistently outperforming other categories of methods in performance.

Recently, an increasing number of studies have turned to RL for improving placement quality. GraphPlace [17] was the first to model macro placement as an RL problem, discretizing the chip canvas into a grid and placing macros step-by-step at specific grid coordinates. However, its reward signal is sparse, providing feedback only after all macros are placed, making learning challenging. Subsequent approaches like DeepPR [18] and PRNet [19] surpassing GraphPlace in performance, but risking violations of non-overlap constraints. To address this, MaskPlace [8] introduces a dense reward mechanism and employs pixel-level visual representations of circuit modules, which fully capture the netlist of thousands of pin connections between macros. It achieves fast placement with zero overlap in large-scale action spaces while maintaining high training efficiency and dense feedback. Further improvements include ChiPFormer [20], which enhances generalization via an offline-trained Decision Transformer, and EfficientPlace [21], which combines global tree search for optimization, delivering high-quality placement results in short timeframes. LaMPlace [22] trains an offline predictor to estimate cross-stage metrics and uses it to generate a pixel-level mask, enabling fast placement decisions based on final chip quality metrics instead of intermediate surrogates.

## III. METHODOLOGY

To address the co-optimization of conflicting wirelength minimization and thermal management objectives, we propose TDPNavigator-Placer, a novel multi-agent reinforcement learning framework that optimize placement based on chiplet's TDP. This approach explicitly delegates these inherently conflicting objectives to specialized agents, each operating under distinct reward mechanisms and environmental constraints within the unified placement paradigm. While single-agent RL struggle with the trade-offs between thermal and wirelength objectives, our multi-agent architecture enables focused optimization through explicit task division.

### A. Framework

As depicted in Fig. 1, there are three main modules in TDPNavigator-Placer, namely TDP navigator, thermal agent and wirelength agent. The TDP Navigator assigns each chiplet to its specialized agent based on whether its TDP exceeds a predefined threshold: high-power chiplets (TDP > threshold) are assigned to the thermal agent for temperature minimization, while low-power chiplets (TDP < threshold) are assigned to the wirelength agent for wirelength optimization.

The RL placement process within each agent is formulated as a Markov Decision Process (MDP), where one chiplet is placed at each step. To enhance thermal efficiency, the placement sequence is determined by TDP, starting with high-power chiplets and followed by low-power ones. This approach is crucial because placing high-TDP chiplets in close proximity is a primary cause of excessive heat. By positioning them first, the algorithm can ensure sufficient spacing is maintained to lower the overall temperature. An episode ends after all chiplets have been placed.

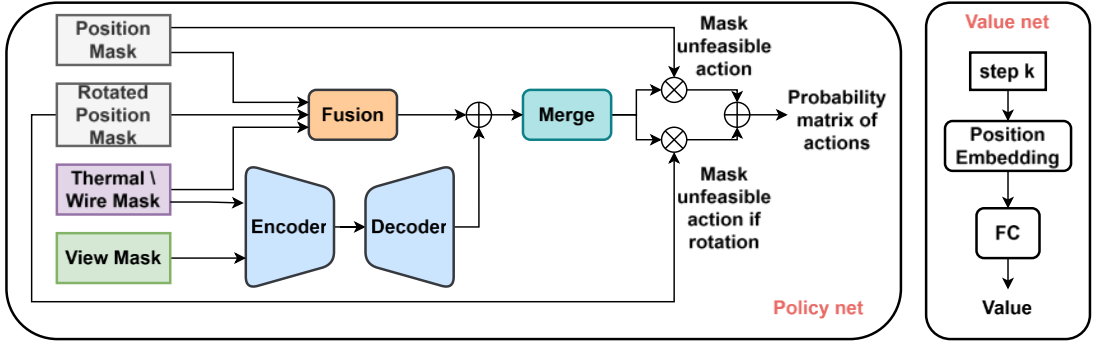

Fig. 2. Overview of thermal-agent and wirelength-agent. All four masks are generated at step k. It shares a similar architecture to MaskPlace [8], with the primary differences being the integration of thermal information into the state and reward, and the doubling of the action space to accommodate rotation. Due to the fixed placement order, the input to the value network is simply the step number rather than the entire state.

*B. Agent*

As shown in Fig. 2, TDPNavigator-Placer treats the canvas as a grid divided into $N \times N$ cells, yielding $2N^2$ discrete actions including position with and without chiplet rotation and employs the proximal policy optimization (PPO) algorithm [12] to train each agent for policy learning. Both agents share three critical input masks: a view mask indicating placed chiplets on the canvas; a position mask identifying feasible locations for the next chiplet; and a rotation position mask specifying feasible locations for the next chiplet rotated by 90 degrees. Additionally, the thermal-agent incorporates a thermal mask as input, which displays the thermal map based on hotspot simulation of placed chiplets. Similarly, the wirelength-agent takes a wire mask [8] as an extra input, indicating the approximate wirelength change resulting from placing the current chiplet at each valid position. All mask values are normalized to the range [-1, 1].

*C. Reward*

The thermal-agent's reward $r_{thermal}$ and the wirelength-agent's reward $r_{wire}$, quantify the reduction in hotspot temperature and wirelength, respectively, after refining the current chiplet placement. Both rewards are normalized to the interval [0, 1]. Specifically, $r_{wire}$ is computed as the difference in wirelength between consecutive steps:

$$r_{wire} = -(WL_k - WL_{k-1}) \quad (1)$$

where $k$ denotes the step index, and $WL_k$ represents the wirelength after step $k$. Similarly, $r_{thermal}$ is defined as:

$$r_{thermal} = -(T_k - T_{k-1}) \quad (2)$$

with $T_k$ being the hotspot temperature after step $k$.

## IV. EXPERIMENTAL EVALUATION

*A. System Configuration*

We evaluate wirelength and temperature of our method on two configurations from the TAP-2.5D environment [5]: Multi-GPU and CPU-DRAM. Table I shows the detailed specifications including chiplet dimensions and TDP profiles. Fig. 3 shows the netlist information. We compare the proposed TDPNavigator-Placer method against three baseline approaches: Simulated Annealing (SA) [5], Bayesian Optimization (BO) and single-agent RL. The single-agent RL framework is similar to our multi-agent version, but differs in that all five masks are fed as input to a single agent. Furthermore, its reward is a weighted sum, with the wirelength and thermal reward weighted at 0.7 and 0.3, respectively. We set the TDP threshold at 80, classifying both the CPU and GPU as high-TDP chiplets.

*B. Experimental Setting*

Our optimization goal is to minimize two competing metrics: wirelength and temperature. These targets naturally conflict with each other: placing chiplets closer together reduces wirelength but increases heat concentration. Because of this trade-off, a simple table isn't effective for comparing our TDPNavigator-Placer with other methods. Instead, we use the Pareto front as our primary performance metric. A Pareto front represents a set of optimal solutions where you can't improve one objective without making the other one worse. In the Pareto-front plot, solutions closer to the bottom-left corner are considered superior, as they represent the best compromises for minimizing both wirelength and temperature.

TABLE I.  BENCHMARK CONFIGURATIONS

|  | Multi-GPU | | | CPU-DRAM | |
|---|---|---|---|---|---|
| Chiplet | CPU | GPU | HBM | CPU | DRAM |
| Width [$mm$] | 12 | 18.2 | 7.75 | 8.25 | 8.75 |
| Height [$mm$] | 12 | 18.2 | 11.87 | 9 | 8.75 |
| TDP[$W$] | 105 | 295 | 20 | 150 | 20 |

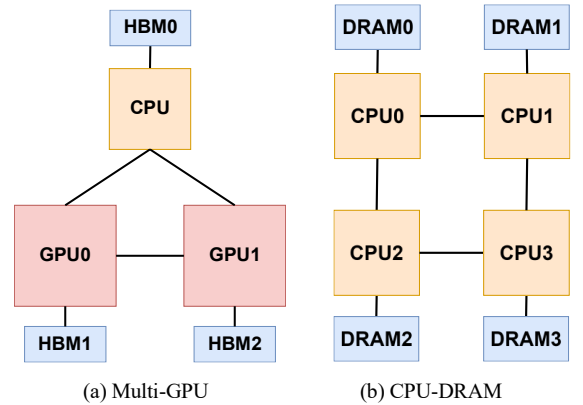

(a) Multi-GPU    (b) CPU-DRAM

Fig. 3. Netlist information

## C. Results

Table II presents the wirelength and temperature results for the proposed TDPNavigator-Placer and the baseline methods. For the Multi-GPU system, it achieves the second-best wirelength performance after TAP-2.5D (SA), while improving temperature by 0.4% compared to the second-best single-agent RL. For the CPU-DRAM design, it achieves a 24% reduction in wirelength and a 3.4% improvement in temperature over the second-best method.

TABLE II. BENCHMARK RESULT

| Method | Multi-GPU | | | |
|---|---|---|---|---|
| | TAP-2.5D [5] | BO | Single-Agent RL | TDPNavigator-Placer |
| Wirelength (mm) ×10⁵ | **0.97** | 1.65 | 1.51 | 1.00 |
| Temperature (°C) | 91.3 | 91.2 | 91.1 | **90.7** |
| Method | CPU-DRAM | | | |
| | TAP-2.5D [5] | BO | Single-Agent RL | TDPNavigator-Placer |
| Wirelength (mm) ×10⁵ | 2.16 | 1.82 | 2.17 | **1.38** |
| Temperature (°C) | 94.9 | 109.8 | 102.8 | **93.7** |

Fig. 4 illustrates the Pareto fronts for wirelength versus temperature across 2 designs in TAP-2.5D system [5]. In both designs, TDPNavigator-Placer achieves lower wirelengths and temperatures, resulting in a more optimal Pareto front. These results highlight TDPNavigator-Placer's advantage in producing thermally efficient and compact placements in both scenarios.

Fig. 5 and Fig. 6 shows the best-performing layouts generated by TDPNavigator-Placer and the corresponding thermal map, highlighting the proposed method's consistent ability to achieve superior thermal and spatial efficiency.

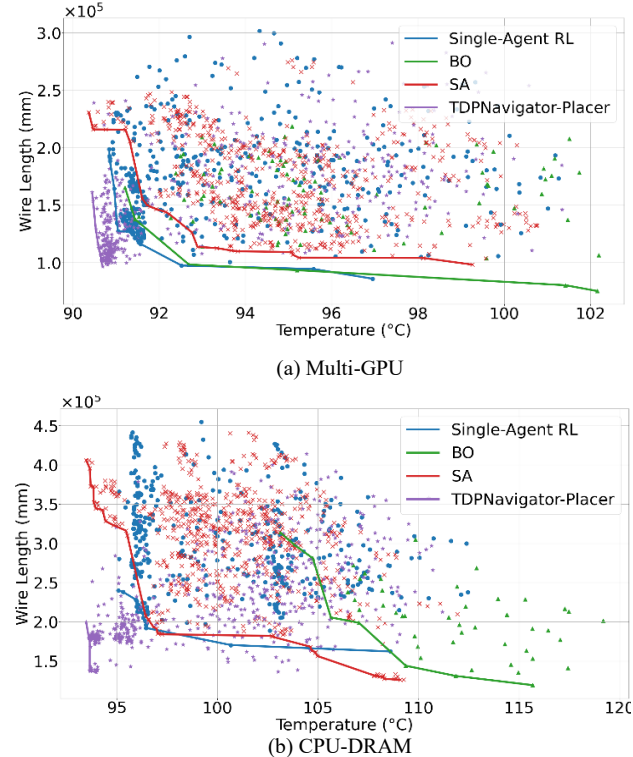

Fig. 4. Pareto fronts across designs

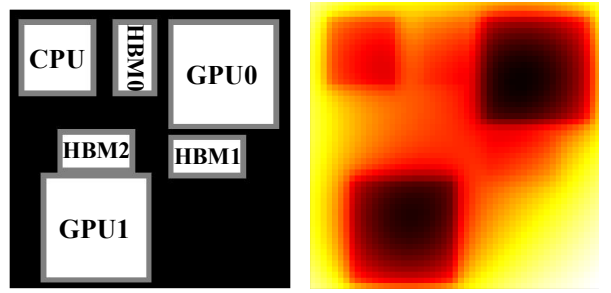

Fig. 5. Chiplet placement layouts generated by TDPNavigator-Placer for Multi-GPU: physical layout (left) and thermal map (right)

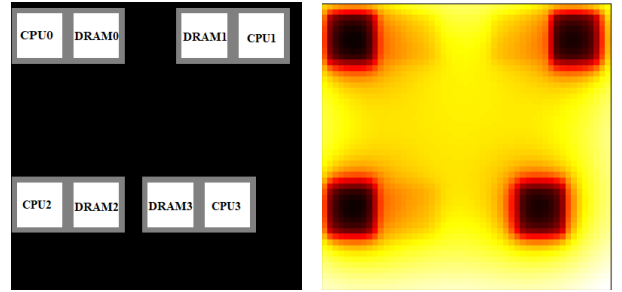

Fig. 6. Chiplet placement layouts generated by TDPNavigator-Placer for CPU-DRAM: physical layout (left) and thermal map (right)

Fig. 7 shows the variation of optimization target and the corresponding cumulative reward in one episode during training. The following is the cumulative reward:

$$R = \sum_{k=1}^{N} r_k \quad (3)$$

where $N$ is the total chiplet number in one design.

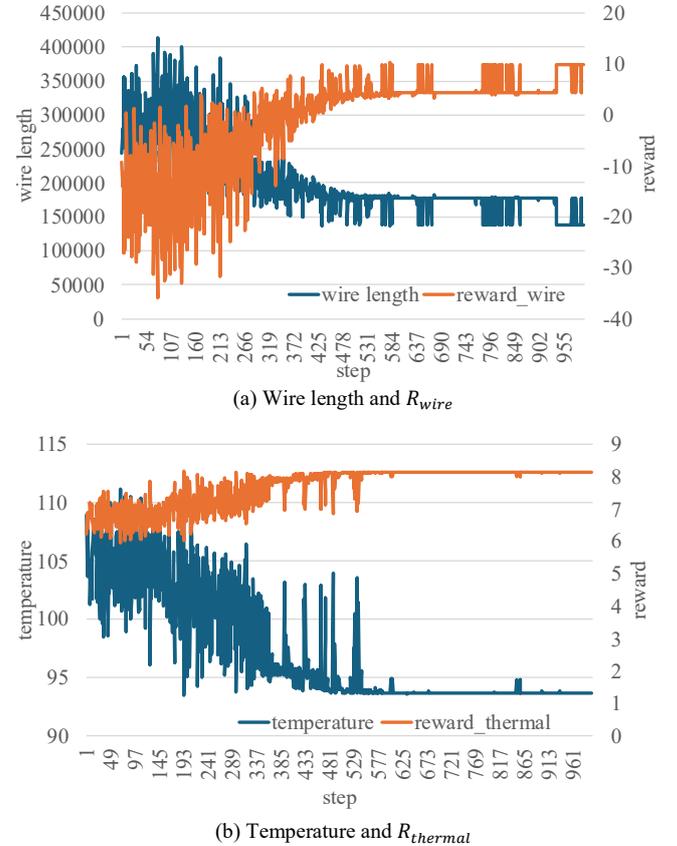

Fig. 7. Compare cumulative reward and performance on CPU-DRAM

As shown in the Fig. 7 (a), the training process begins with an exploration phase, where the cumulative reward

$R_{wire}$ fluctuates around -20. As training progresses, $R_{wire}$ steadily increases, which corresponds to a decrease in the overall wirelength. Eventually, the reward curve converges, indicating that the model has successfully learned an effective policy for optimizing the chiplet layout. Similarly, as shown in Fig. 7 (b), the $R_{thermal}$ starts at a relatively low value during exploration and progressively decreases throughout training, corresponding to the thermal policy becoming more deterministic and exploiting learned optimal actions.

## V. CONCLUSION

In this paper, we addresses the critical challenge of thermal-wirelength trade-offs in 2.5D IC placement amid growing demands for cost-effective, scalable computing systems. While prior approaches fail to resolve the conflict between wirelength minimization and thermal management, we propose TDPNavigator-Placer, a novel MARL framework that pioneers explicit task decomposition for concurrent optimization. Specifically, our solution uses a TDP Navigator that dynamically assigns chiplets to specialized agents based on their thermal design power (TDP). Subsequently, wirelength and thermal agents operates under distinct reward mechanisms and environmental constraints within a unified placement paradigm. This approach facilitates effective thermal-aware co-optimization in advanced packaging, paving the way for sustainable, high-performance 2.5D integration.

ACKNOWLEDGMENT

This research is supported by the Agency for Science, Technology and Research (A*STAR) under its MTC Programmatic Funds (Grant No. M23M3b0064).